%% file: main.tex
\documentclass[10pt,twocolumn,letterpaper]{article}

\usepackage[pagenumbers]{cvpr}      % To produce the REVIEW version

\definecolor{cvprblue}{rgb}{0.21,0.49,0.74}
\usepackage[pagebackref,breaklinks,colorlinks,allcolors=cvprblue]{hyperref}
\usepackage{soul}
\usepackage{makecell}
\usepackage{multirow}
\usepackage{amssymb} % 添加 amssymb 以使用 \checkmark
\usepackage{threeparttable} 
\usepackage{float}
\usepackage{graphicx}
\usepackage{epstopdf}

\def\paperID{11132} % *** Enter the Paper ID here
\def\confName{CVPR}
\def\confYear{2025}

\title{DiFiC: Your Diffusion Model Holds the Secret to Fine-Grained Clustering}

\author{Ruohong Yang\textsuperscript{1}, Peng Hu\textsuperscript{1}, Xi Peng\textsuperscript{1}, Xiting Liu\textsuperscript{2*}, Yunfan Li\textsuperscript{1*}\\
\textsuperscript{1} College of Computer Science, Sichuan University.\\
\textsuperscript{2} Georgia Insitute of Technology.\\
{\tt\small \{ruohong.yrh,  penghu.ml, pengx.gm, yunfanli.gm\}@gmail.com,}\\{\tt\small lxt670@126.com}
}
\begin{document}
\maketitle

\begin{abstract}
Fine-grained clustering is a practical yet challenging task, whose essence lies in capturing the subtle differences between instances of different classes. Such subtle differences can be easily disrupted by data augmentation or be overwhelmed by redundant information in data, leading to significant performance degradation for existing clustering methods. In this work, we introduce DiFiC a fine-grained clustering method building upon the conditional diffusion model. Distinct from existing works that focus on extracting discriminative features from images, DiFiC resorts to deducing the textual conditions used for image generation. To distill more precise and clustering-favorable object semantics, DiFiC further regularizes the diffusion target and guides the distillation process utilizing neighborhood similarity. Extensive experiments demonstrate that DiFiC outperforms both state-of-the-art discriminative and generative clustering methods on four fine-grained image clustering benchmarks. We hope the success of DiFiC will inspire future research to unlock the potential of diffusion models in tasks beyond generation. The code will be released.
\end{abstract}

\section{Introduction}

\begin{figure}[t]
  \centering
   \includegraphics[width=0.95\linewidth]{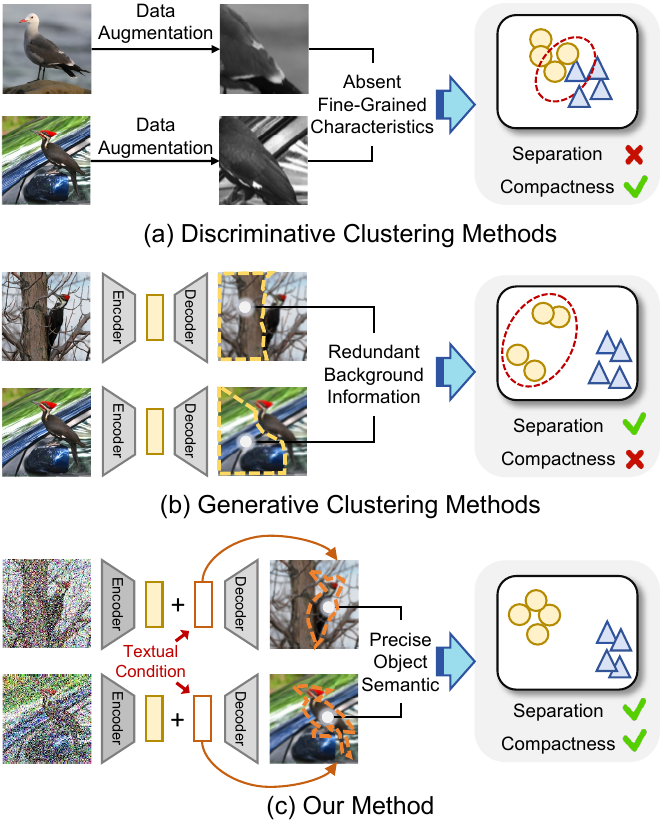}
   \caption{Our key idea. Existing clustering methods struggle to capture the subtle signals to distinguish fine-grained data. Specifically, (a) discriminative clustering methods heavily rely on data augmentation, which may disrupt subtle semantic differences, leading to inferior between-cluster separation. (b) Generative clustering methods infer latent features for reconstructing all pixels, in which redundant background information may overwhelm fine-grained semantics, resulting in suboptimal within-cluster compactness. (c) Instead of directly learning image features, our method resorts to deducing the textual conditions used for image generation, overcoming the semantic absence or redundancy problem suffered by previous works.
   }\vspace{-2mm}
   \label{fig:motivate}
\end{figure}
   
Clustering aims at grouping instances into distinct clusters based on semantic similarity. In recent years, thanks to the powerful feature extraction ability of deep neural networks, deep clustering methods \cite{survey,deepclu} have witnessed great success in real-world applications such as anomaly detection \cite{anomaly}, community discovery \cite{community}, and bioinformatics \cite{bio1}. Based on the clustering paradigm, existing deep clustering studies could be broadly categorized into discriminative and generative approaches. In brief, discriminative clustering methods \cite{dis1,ccdis5,secudis8} focus on optimizing the cluster boundaries, by pulling within-cluster instances together while pushing between-cluster instances apart. In comparison, generative clustering methods \cite{clustergan,infogangen1,c3gangen4} assume that instances with different semantics follow distinct data distributions, and group instances generated from similar distributions together to achieve clustering.

Though having achieved promising results, most existing deep clustering methods are designed for coarse-grained datasets. Little attention has been paid to the more challenging fine-grained clustering task, which is essential for real-world applications. In fact, both current discriminative and generative methods struggle to capture the fine-grained, subtle differences between images. Specifically, as shown in \cref{fig:motivate}(a), most discriminative clustering methods heavily rely on the data augmentation strategy. However, fine-grained characteristics, such as the birds' crown style, would likely be disrupted by augmentations like color jittering and random cropping, leading to inferior between-cluster separation. For generative clustering methods, since the latent features are expected to reconstruct all image pixels, they would be inevitability contaminated by the redundant background information, which overwhelms the fine-grained semantics. For example, as shown in \cref{fig:motivate}(b), images of birds from the same species may exhibit distinct latent features given the significant variations in their backgrounds, resulting in inferior within-cluster compactness.

In this work, we introduce DiFiC, a novel fine-grained clustering method that overcomes the aforementioned limitations as illustrated in \cref{fig:motivate}(c). On the one hand, compared with discriminative clustering methods, DiFiC does not require risky data augmentations, ensuring the subtle characteristics are well preserved. On the other hand, different from existing generative clustering methods that focus on mining image latent features, DiFiC resorts to distilling the textual conditions used for image generation, which contains more compact semantics. Moreover, instead of generating the entire image, we propose to estimate the object location based on attention maps and regularize the generation target accordingly. This effectively filters out irrelevant background information, allowing the distilled textual semantics to focus on the main object. Finally, we incorporate neighborhood similarly to align the semantic distillation process with clustering objectives. On four fine-grained image clustering datasets, DiFiC enjoys superior between-cluster separation and within-cluster compactness, achieving state-of-the-art performance compared with existing baselines.

The major contributions of this work could be summarized as follows:
\begin{itemize}
\item We tackle an important but less-explored problem, fine-grained clustering, which most existing clustering methods struggle to handle. In brief, the key to fine-grained clustering lies in capturing the subtle differences between classes. However, such a subtle signal would be disrupted by the data augmentations in discriminative methods or be overwhelmed by the redundant background information in generative methods.

\item We propose a fine-grained clustering method DiFiC, building upon the conditional diffusion model. Instead of focusing on extracting representative features from images, we resort to deducing the textual conditions used for image generation, which encapsulate more compact semantics. To focus the semantics on objects, we design an attention-based mask to regularize the diffusion target. Additionally, we leverage the neighborhood similarity to align the semantics deducing with clustering objectives.

\item Experiments on four fine-grained image clustering datasets demonstrate the effectiveness and superiority of DiFiC compared with existing methods. Moreover, the success of DiFiC offers a fresh perspective on applying generative models to discriminative tasks, such as classification and clustering.
\end{itemize}

\section{Related Work}
In this section, we briefly review two related topics, namely, deep clustering and conditional diffusion.

\subsection{Deep Clustering}
Deep clustering methods \cite{deep1, deep2, deep3, deep4, survey}, powered by the feature extraction capability of neural networks, have made significant advancements in recent years. Existing deep clustering studies could be broadly categorized into two main paradigms, discriminative \cite{dis2, iicdis4, gccdis7, TCLdis6} and generative \cite{infogangen1, mixnmatch, c3gangen4}. Specifically, discriminative clustering methods optimize the cluster boundaries by pulling within-cluster instances together, while pushing between-cluster instances apart. As a representative, contrastive clustering methods \cite{xcc, gccdis7, ccdis5, TCLdis6, secudis8} employ data augmentation to construct positive and negative sample pairs.
% (\textit{i.e.}, positive pairs) (\textit{i.e.}, negative pairs)
% For example, IIC \cite{iicdis4} maximizes the mutual information between each image and its augmentations. CC \cite{ccdis5} further performs contrastive learning at the cluster level. Recent works \cite{TCLdis6, secudis8} leverage pseudo-labeling to guide the pair construction.
Another line of research is generative clustering, based on the assumption that instances from different clusters follow distinct distributions. Leveraging such distribution prior, generative clustering methods learn the underlying data distributions through models such as variational autoencoders \cite{vade} and generative adversarial networks \cite{clustergan, infogangen1, c3gangen4}, grouping instances with similar distributions to achieve clustering.

Though having achieved promising performance on coarse-grained data, most existing methods struggle with the more challenging fine-grained clustering task. For discriminative clustering methods, data augmentation can obscure subtle differences between images, undermining fine-grained distinction. Meanwhile, for generative clustering methods, the learned latent distributions would be contaminated by redundant background information, which overwhelms the fine-grained semantics. To address these limitations, we propose DiFiC, a novel deep clustering method tailored for fine-grained data. Unlike existing approaches that emphasize extracting representative features from images, DiFiC instead infers the textual conditions used for image generation, enabling a more accurate capture of fine-grained semantics.

\begin{figure*}
  \centering
   \includegraphics[width=0.96\linewidth]{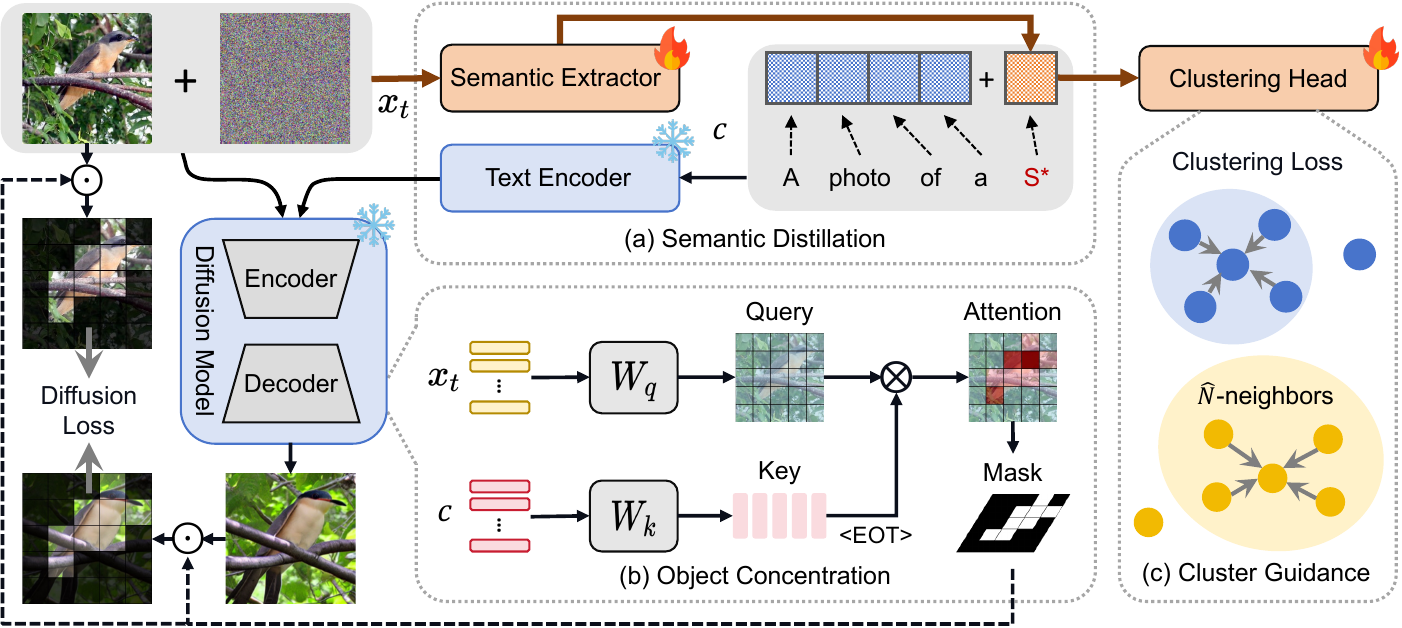}
   \caption{Overview of the proposed DiFiC, a fine-grained image clustering method built upon a pre-trained text-to-image diffusion model, which consists of three main modules: (a) \textbf{Semantic Distillation Module}: for each noisy image $x_t$, DiFiC first extracts its semantical proxy word $S^*$ using a semantic extractor, which is then concatenated with a prompt to form the textual condition $c$ for image generation. By requiring the diffusion model to restore the original image, DiFiC distills the image semantics into the proxy word. (b) \textbf{Object Concentration Module}: instead of restoring the full image, DiFiC computes the object mask based on attention maps, which is applied to both the original and generated images when calculating the diffusion loss. As a result, the distilled semantics would center on the main object. (c) \textbf{Cluster Guidance Module}: given proxy word embeddings, DiFiC introduces a clustering head to group images based on neighborhood similarity. The clustering loss simultaneously optimizes the semantic extractor and clustering head, guiding the distillation for producing more compact semantics. In the figure, the bold brown arrows denote the data flow to achieve clustering.}
   \label{fig:over}
\end{figure*}

\subsection{Conditional Diffusion}

By progressively destructing an image with noise and learning the reverse process, diffusion models have achieved promising results in image generation \cite{dm1, dm2, dm3, dm4}. To control the semantics of the generated images, conditional diffusion models integrate textual conditions during the diffusion process, exhibiting strong generative capabilities and versatility in tasks such as image editing \cite{imageedit1, imageedit2, imageedit3} and image restoration \cite{imagerestore1, imagerestore2}. For instance, Glide \cite{glide} incorporates textual features directly into transformer blocks for conditional generation and editing. Stable Diffusion \cite{sd} performs text-to-image diffusion in the latent space, significantly improving the efficiency and scalability of conditional diffusion models. Moreover, although these models are primarily designed for generative tasks, researchers have also investigated their potential for representation learning \cite{rep1}, extending their application to discriminative downstream tasks such as classification \cite{class1, beat} and segmentation \cite{seg1, seg2}.

However, as diffusion models are designed to generate all image pixels, their latent features inevitably contain extensive low-level information, limiting their discriminative capability. In this work, rather than focusing on learning image features, we propose distilling semantic information from the textual condition used for image generation, which proves highly effective for fine-grained image clustering. To the best of our knowledge, this could be the first work to leverage diffusion models for clustering, and we hope it will inspire future research to unlock the potential of diffusion models in tasks beyond generation.

\section{Method}

In this section, we present DiFiC, a fine-grained image clustering method built upon a pre-trained text-to-image diffusion model. As illustrated in \cref{fig:over}, DiFiC consists of three modules. First, the \textbf{semantic distillation module} deduces the textual condition used by the diffusion model in image generation. Then, the \textbf{object concentration module} regularizes the diffusion target with object masks derived from attention maps, helping the textual condition focus on object semantics. Lastly, the \textbf{cluster guidance module} aligns the semantic distillation process with clustering objectives, enabling effective image grouping based on the distilled semantics. The three modules are elaborated below.

\subsection{Semantic Distillation} \label{sec:Diffusion Inversion}

A conditional diffusion model consists of a UNet $\boldsymbol{\epsilon}_\theta(\cdot)$ and a text encoder $\tau_\theta(\cdot)$. Given the input image $x_0$ and text prompt $c$, the model iteratively generates an image by predicting and removing noise at each timestep $t$, with the following objective function, \textit{i.e.},
\begin{align}
\mathcal{L}_{{DM}} = \mathbb{E}_{t, x_0, {\epsilon}}\left[\left\|{\epsilon}-\boldsymbol{\epsilon}_\theta\left(x_t, t, \tau_\theta(c)\right)\right\|^2\right], \label{eq:ldiff} \\
{x}_t=\sqrt{{\alpha}_t} {x_0} +\sqrt{1-{\alpha}_t} {\epsilon}, {\epsilon} \sim \mathcal{N}(0, \mathbf{I}),~~ \label{eq:xt}
\end{align}
where $x_t$ represents the noisy image generated by sampling Gaussian noise ${\epsilon}$ on ${x_0}$, and ${\alpha}_t$ controls the noise schedule at timestep $t$ \cite{ddim}. 

Unlike the common use of \cref{eq:ldiff} to train the models $\boldsymbol{\epsilon}_\theta(\cdot)$ and $\tau_\theta(\cdot)$, we propose to deduce the text prompt $c$ used to generate the image $x_0$ from a pre-trained diffusion model. The deduced text prompt encapsulates the image semantics, which could thus be used for clustering. To achieve semantic distillation, the most straightforward solution is to deduce one text prompt $c$ for each image. However, such an implementation has two inherent limitations. On the one hand, it is daunting to handle large datasets, as the time-consuming timestep sampling process must be repeated for every image. On the other hand, the distillation process is independent for each image, hindering the potential collaboration between different images.

Therefore, instead of directly optimizing the text prompt $c$, we introduce a shared semantic extractor $f(\cdot)$, which maps the image feature ${z}_{\hat{t}}^b$ to the proxy word $S^*$ in the text prompt $c$ (\textit{e.g.}, $c =$ ``A photo of a $S^*$''). Here, the image feature ${z}_{\hat{t}}^b$ corresponds to the mid-layer representation from the $b$-th block of the UNet (denoted as $\boldsymbol{\epsilon}_\theta^b$), namely,
\begin{equation}
{z}_{\hat{t}}^b=\boldsymbol{\epsilon}_\theta^b\left({x}_{\hat{t}}, \hat{t}, \tau_\theta(\hat{c})\right),
\label{eq:hid}
\end{equation}
where $\hat{t}=150$ is the timestep at which the image feature is computed, and $\hat{c}$ refers to the empty text prompt. Notably, to prevent inconsistent Gaussian noises from disturbing the image features, we sample the same noise $\epsilon$ at timestep $\hat{t}$ for all images when computing ${z}_{\hat{t}}^b$. We set $b=19$ in all the experiments.

Consequently, to perform semantic distillation, we rewrite the objective function in \cref{eq:ldiff} as follows,
\begin{align}
\mathcal{L}_{{SD}} =\mathbb{E}_{t, {x_0}, {\epsilon}}\left[\left\|{\epsilon}-\boldsymbol{\epsilon}_\theta\left({x}_t, t, \tau_\theta\left(S^*\right)\right)\right\|^2\right], \label{eq:lsd} \\
S^* = f({z}_{\hat{t}}^b),\label{eq:s}~~~~~~~~~~~~~~~~~~~~~~~~~~ 
\end{align}
where text prompts such as ``A photo of a'' are omitted for simplicity. By minimizing \cref{eq:lsd}, the semantic extractor $f$ is able to distill the image semantic into the proxy word $S^*$. Since $f$ is shared across different images, it requires less time to converge compared with independently distilling textual conditions for each image.

Furthermore, during the denoising process, diffusion models tend to first restore high-level semantics (corresponding to large timesteps) and then fill in low-level details (corresponding to small timesteps). Since fine-grained clustering aims at discriminating images with subtle differences, we propose the following weighted timestep sampling scheme during semantic distillation, \textit{i.e.},
\begin{equation}
    w(t) = 1 + cos(\frac{\pi t}{T}),
    \label{eq:wt}
\end{equation}
where $w(t)$ corresponds to the probability of timestep $t$ being sampled from the total diffusion timesteps $T$. By emphasizing small diffusion timesteps, the distilled proxy word $S^*$ could better capture fine-grained details, benefiting the subsequent clustering process.

\subsection{Object Concentration} \label{sec:Attention Mask}
Since diffusion models are trained to restore all pixels in an image, the distilled textual conditions would inevitably contain redundant background information, interfering with clustering based on foreground objects. To mitigate this, we introduce an object-centered mask to regularize the diffusion target, ensuring the diffusion loss in \cref{eq:lsd} is only computed on the main object.

To locate the main object in images, we resort to attention maps which possess good properties of semantic segmentation. Specifically, given an image $x_0$ with height $H$ and width $W$ and a text prompt $\hat{c}$, the cross-attention matrix in the $i$-th UNet block is computed between the image query $Q_i$ and the text key $K_i$ of $[\mathrm{EOT}]$ token, namely,
\begin{align}
A(Q_i,K_i)=\operatorname{softmax}\left(\frac{Q_i K_i^{\top}}{\sqrt{d}}\right) \in \mathbb{R}^{\left\lceil\frac{H}{r_i}\right\rceil \times\left\lceil\frac{W}{r_i}\right\rceil}, \label{eq:atten} \\
Q_i=W_q^i\left(z_{\tilde{t}}^i\right), \quad K_i=W_k^i\left(\tau_\theta(\hat{c})\right)[\mathrm{EOT}],~~~~
\end{align}
where $W_q^i, W_k^i$ denote the weight matrices projecting image feature $z_{\tilde{t}}^i$ and text embedding $\tau_\theta(\hat{c})$ into query and key, respectively; $z_{\tilde{t}}^i$ represents the representation at timestep $\tilde{t}=50$; $r_i$ is the downsampling factor; and $d$ is the feature dimension. In practice, we average the cross-attention matrices from all attention heads within each UNet block.

To aggregate the attention matrices across all $L$ blocks, we resize all $A(Q_i, K_i), i \in [1, L]$ to match the shape of the added Gaussian noise $\epsilon$, and compute the block-wise average attention map via
\begin{equation}
\bar{A}=\frac{1}{L} \sum_{i=1}^L A(Q_i,K_i).
\end{equation}
The average attention map $\bar{A}$ carries the layout information of the main object. To compute a bipartite mask to distinguish foreground and background, we model the attention value distribution using a two-component Gaussian Mixture Model, namely,
\begin{equation}
P(\bar{A})=\gamma_1 \phi\left(\bar{A} \mid \mu_1, \sigma_1^2\right)+\gamma_2 \phi\left(\bar{A} \mid \mu_2, \sigma_2^2\right),
\end{equation}
where $\gamma_1$ and $\gamma_2$ are the mixture coefficients, $\phi\left(\bar{A} \mid \mu_k, \sigma_k^2\right)$ represents the probability density of the components with mean $\mu_k$ and variance $\sigma_k^2$ ($k=1,2$). The bipartite mask $\mathbf{m}$ is then calculated by
\begin{equation}
\mathbf{m}_{ij}= \begin{cases}1, & \text { if } \bar{A}_{ij}>\frac{1}{2}\left({\mu_1+\mu_2}\right), \\ 0, & \text { otherwise. }\end{cases}
\end{equation}

To concentrate on the main object during semantic distillation, we rewrite \cref{eq:lsd} by applying the bipartite mask $\mathbf{m}$ as follows,
\begin{align}
\mathcal{L}_{{SD}}^{M} = & \mathbb{E}_{t \sim w(t), {x_0}, {\epsilon}} \Bigg[ \Big\| {\epsilon} \odot {\mathbf{m}} -  \nonumber \\
& \vspace{-1em} \boldsymbol{\epsilon}_\theta\left({x}_t, t, \tau_\theta\left(f({z}_{\hat{t}}^b)\right)\right) \odot {\mathbf{m}} \Big\|^2 \Bigg],
\label{eq:msd}
\end{align}
where $\odot$ denotes Hadamard product.

\subsection{Cluster Guidance} \label{sec:Neighborhood Clustering}
Thanks to object-concentrated semantic distillation, the proxy words contain rich semantics of the main objects in images. To further align the distilled image semantics with clustering objectives, we introduce a clustering head to group the proxy words based on neighborhood similarity.

To elaborate, given a set of proxy words $\{S_i^*\}_{i=1}^N$, the clustering head $g(\cdot)$ predicts soft cluster assignment $p_i \in \mathbb{R}^{C}$ for each word $S_i^*$, where ${C}$ denotes the cluster number. Let $\mathcal{N}(S_i^*)$ be the indices of $\hat{N}=10$ nearest neighbors of $S_i^*$. We encourage consistent clustering assignments between $S_i^*$ and its neighboring proxy words, by optimizing the following clustering objective, \textit{i.e.},
\begin{align}
\mathcal{L}_{{NB}}=-\frac{1}{N}\log \sum_{i=1}^N p_i^{\top} p_j,~~j \sim \mathcal{N}(S_i^*), \\
p_i = g(S_i^*),~~p_j = g(S_j^*), \label{eq:p} ~~~~~~~~~
\end{align}
where $p_i, p_j$ refer to the cluster assignment of $S_i^*$ and one randomly selected neighbor $S_j^*$.

Moreover, to prevent $g(\cdot)$ from assigning most samples to only a few clusters, we introduce an entropy regularization term, namely,
\begin{equation}
\mathcal{L}_{ {EN}}=-\sum_{i=1}^{C} \bar{p}_i \log (\bar{p}_i),~~\bar{p} = \frac{1}{N} \sum_{i=1}^N p_{i} \in \mathbb{R}^C.\label{eq:en}
\end{equation}

To accurately reflect the overall cluster assignment distribution, the number of samples $N$ within each batch should exceed the cluster number $C$. However, due to the extensive memory requirements of diffusion models, the batch size $N$ is often much smaller than $C$. Consequently, the entropy regularization term cannot effectively reflect the overall cluster assignment distribution, leading to inferior performance. As a solution, we incorporate a memory bank of length $U=512$ to store the historical cluster assignments, which participate in computing $\bar{p}$ in \cref{eq:en}. The total clustering loss lies in the form of
\begin{equation}
\mathcal{L}_{CG}=\mathcal{L}_{{NB}}-\lambda \cdot \mathcal{L}_{ {EN}},\label{eq:cg}
\end{equation}
where $\lambda=5 \times\left(\frac{U+N}{N}\right)$ weights the strength of entropy regularization. The above loss optimizes not only the clustering head $g(\cdot)$ to achieve clustering, but also the semantic extractor $f(\cdot)$ to guide the semantic distillation.

Combining \cref{eq:msd} and \cref{eq:cg}, we arrive at the overall objective function of DiFiC, \textit{i.e.},
\begin{equation}
\mathcal{L}_{\mathrm{DiFiC}}=\mathcal{L}_{{SD}}^{M}+\mathcal{L}_{{CG}}.
\end{equation}

After training, for each image $x_0$, DiFiC first computes the noisy image $x_{\hat{t}}$ by \cref{eq:xt}, and then extracts the semantical proxy word $S^*$ through \cref{eq:hid,eq:s}. Finally, DiFiC predicts the soft clustering assignment of $x_0$ via \cref{eq:p}.

\section{Experiments}

To evaluate the proposed DiFiC, we apply it to four classic fine-grained image clustering datasets. Then, we conduct a series of performance comparisons, ablation studies, and parameter analyses to demonstrate the effectiveness of DiFiC.

\subsection{Datasets and Evaluation Metrics}

Four classic fine-grained image clustering datasets are used for evaluation, including CUB~\cite{cub}, Car~\cite{car}, Dog~\cite{dog}, and Flower~\cite{flower}. The brief information of these datasets is summarized in \cref{tab:dataset}. Notably, we train clustering models on the training split and evaluate them on the test split. In line with the previous studies~\cite{finegangen2, mixnmatch, onegangen3, c3gangen4}, the training splits for CUB and Flower include the entire dataset, due to the small number of training images.

We adopt two widely used metrics, clustering accuracy (ACC) and normalized mutual information (NMI), to evaluate the clustering performance. Higher values for both metrics indicate better results.

\begin{table}[!h]
  \caption{A summary of datasets used for evaluation.}  
  \label{tab:dataset}
  \centering
  \setlength{\tabcolsep}{3.4pt}
  \begin{tabular}{@{}lccccc@{}}  % 四列：左对齐和三列居中对齐
    \toprule
    Dataset & Training Split & Test Split & Training & Test & Classes \\  % 修正标题
    \midrule
    CUB &  Train $+$ Test & Test & 5,994 & 5,794 & 200 \\
    Car &  Train & Test &  8,144 & 8,041 & 196 \\
    Dog &  Train & Test &  12,000 & 8,580 & 120 \\
    Flower &  Train $+$ Test & Test &  2,040 & 6,149 & 102 \\
    \bottomrule
  \end{tabular}
\end{table}

\begin{table*}[!t]
  \caption{Performance on four fine-grained image clustering benchmarks, with the best results denoted in bold. $^{\dagger}$ denotes performing $k$-means on features. $^{\ddagger}$ refers to the semi-supervised variant using bounding boxes. $^{\ast}$ means over-clustering results.}
  \label{tab:sota}
  \centering
  \resizebox{\textwidth}{!}{
  \begin{tabular}{clccccccccc}
    \toprule
    & \multirow{2}{*}{Method}   & \multirow{2}{*}{Publication} &\multicolumn{4}{c}{ACC $\uparrow$} & \multicolumn{4}{c}{NMI $\uparrow$} \\
    \cmidrule(lr){4-7} \cmidrule(lr){8-11}
    \multicolumn{3}{c}{}   
     & CUB & Car & Dog & Flower & CUB & Car & Dog & Flower \\
    \midrule
    \multirow{5}{*}{\begin{tabular}[c]{@{}c@{}}Discriminative \\Methods\\\end{tabular}} & IIC \cite{iicdis4} & ICCV'2019  & 7.4 & 4.9 & 5.0 & 8.7 & 0.36 & 0.27 & 0.18 & 0.24 \\
    & SimCLR$^{\dagger}$ \cite{simclrcl1} & ICML'2020 & 8.4 & 6.7 & 6.8 & 12.5 & 0.40 & 0.33 & 0.19 & 0.29 \\
    & MoCo$^{\dagger}$ \cite{mococl2} & CVPR'2020 & 10.2 & 8.0 & 11.5 & 51.5 & 0.37 & 0.37 & 0.31 & 0.68 \\
    & SCAN \cite{scanpseudo3} & ECCV'2020 & 11.9 & 8.8 & 12.3 & 56.5 & 0.45 & 0.38 & 0.35 & 0.77 \\
    & SeCu \cite{secudis8}& ICCV'2023 & 15.4 & 10.0 & 16.7 & 68.6 & 0.48 & 0.39 & 0.38 & 0.84\\
    \midrule
    \multirow{10}{*}{\begin{tabular}[c]{@{}c@{}}Generative \\Methods\\\end{tabular}}
    & InfoGAN \cite{infogangen1}& NeurIPS'2016  & 8.6 & 6.5 & 6.4 & 23.2 & 0.39 & 0.31 & 0.21 & 0.44 \\
    & FineGAN \cite{finegangen2} & CVPR'2019 & 6.9 & 6.8 & 6.0 & 8.1 & 0.37 & 0.33 & 0.22 & 0.24 \\
     & FineGAN$^{\ddagger}$ \cite{finegangen2} & CVPR'2019 & 12.6 & 7.8 & 7.9 & - & 0.40 & 0.35 & 0.23 & - \\
    & MixNMatch \cite{mixnmatch} & CVPR'2020 & 10.2 & 7.3 & 10.3 & 39.0 & 0.41 & 0.34 & 0.30 & 0.57 \\
    & MixNMatch$^{\ddagger}$ \cite{mixnmatch} &CVPR'2020 & 13.6 & 7.9 & 8.9 & - & 0.42 & 0.36 & 0.32 & - \\
    & OneGAN$^{\ddagger}$ \cite{onegangen3} & ECCV'2020 & 10.1 & 6.0 & 7.3 & - & 0.39 & 0.27 & 0.21 & - \\
    & Stable Diffusion$^{\dagger}$ \cite{sd} & CVPR'2022  & 7.6 & 9.1  &  5.4 & 13.1 & 0.34 & 0.37 & 0.18 & 0.34\\
 %    & DiFiC$^{\ast}$ (Ours) & -
 % & 27.1 &  33.2 & 13.5  & 43.2 &  0.60 & 0.64  & \textbf{0.42}  & 0.77 \\
    & C3-GAN \cite{c3gangen4}& ICLR'2022 & 22.7 & 8.3 & 11.8 & 55.6 &  0.50 &  0.33 &  0.30 & 0.72 \\
    & C3-GAN$^{\ast}$ \cite{c3gangen4} & ICLR'2022 & 19.7 & 7.2 & 12.5 & 51.2 & 0.53 & 0.41 & 0.36 & 0.67 \\
    & \textbf{DiFiC (Ours)} & -
 & \textbf{31.7} & \textbf{47.2}  &  \textbf{17.2} & \textbf{72.9} &  \textbf{0.61} & \textbf{0.68}   & \textbf{0.39} & \textbf{0.88}   \\
    \bottomrule
  \end{tabular}}
\end{table*}

\subsection{Implementation Details}

Without loss of generality, we apply the proposed DiFiC on the Stable Diffusion v1-5 model~\cite{sd}, which offers a good balance between generation speed and quality. All images are generated at a resolution of $512 \times 512$. The semantic extractor $f(\cdot)$ consists of two blocks, each featuring a $2 \times 2$ convolution and a $2 \times 2$ max-pooling, a $2 \times 2$ adaptive average-pooling, followed by a linear projection layer with dimensions of $3072$-$768$. The clustering head $g(\cdot)$ is a two-layer MLP with dimensions of $768$-$768$-$C$, where $C$ corresponds to the cluster number. We train the model for 250 epochs under a batch size of $32$, using the AdamW optimizer with a learning rate of $2e-4$. In the first 100 training epochs, we use only $\mathcal{L}_{{SD}}^{M}$ in \cref{eq:msd} to warm up the semantic extractor $f(\cdot)$. The $\hat{N}$-nearest neighbors are computed at the end of each training epoch. All experiments are conducted on four NVIDIA RTX 3090 GPUs.

\subsection{Main Results}

\begin{figure*}[t]
  \centering
  \begin{subfigure}{0.24\linewidth}
    \centering
    \includegraphics[width=\linewidth]{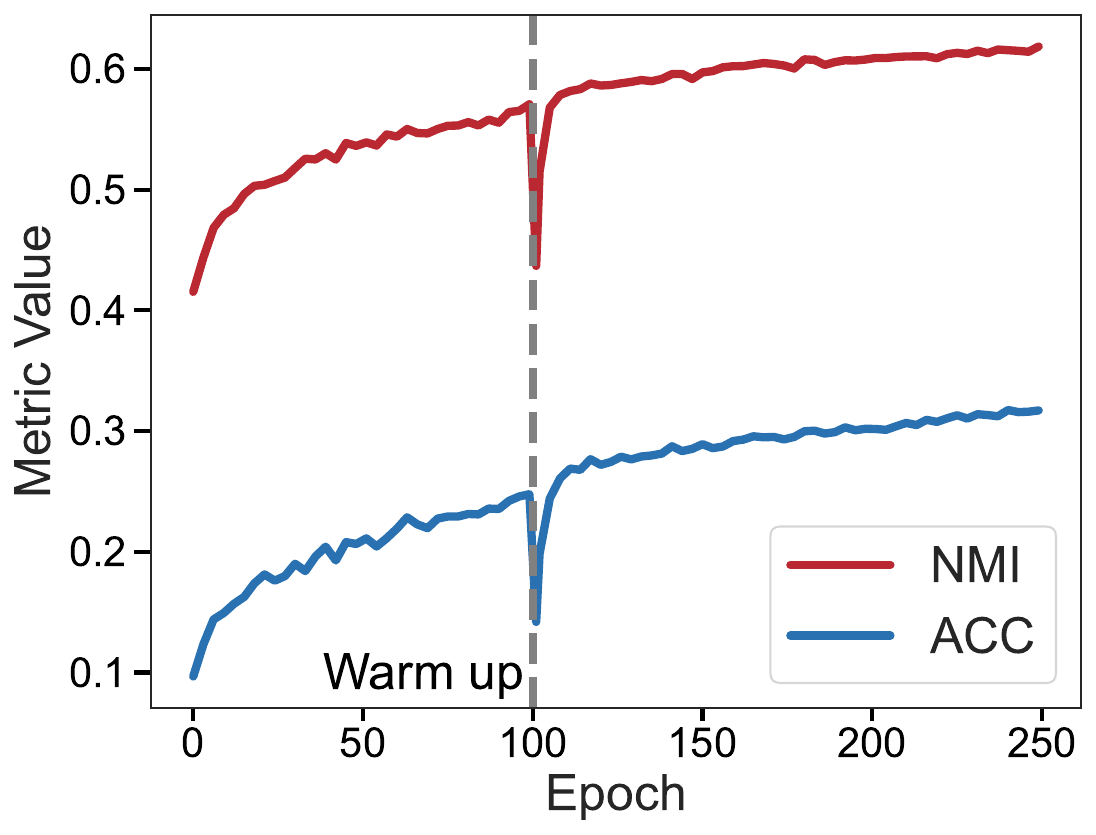} % Replace with actual image path
    \caption{Performance on CUB}
    \label{fig:acc-a}
  \end{subfigure}
  \begin{subfigure}{0.24\linewidth}
    \centering
    \includegraphics[width=\linewidth]{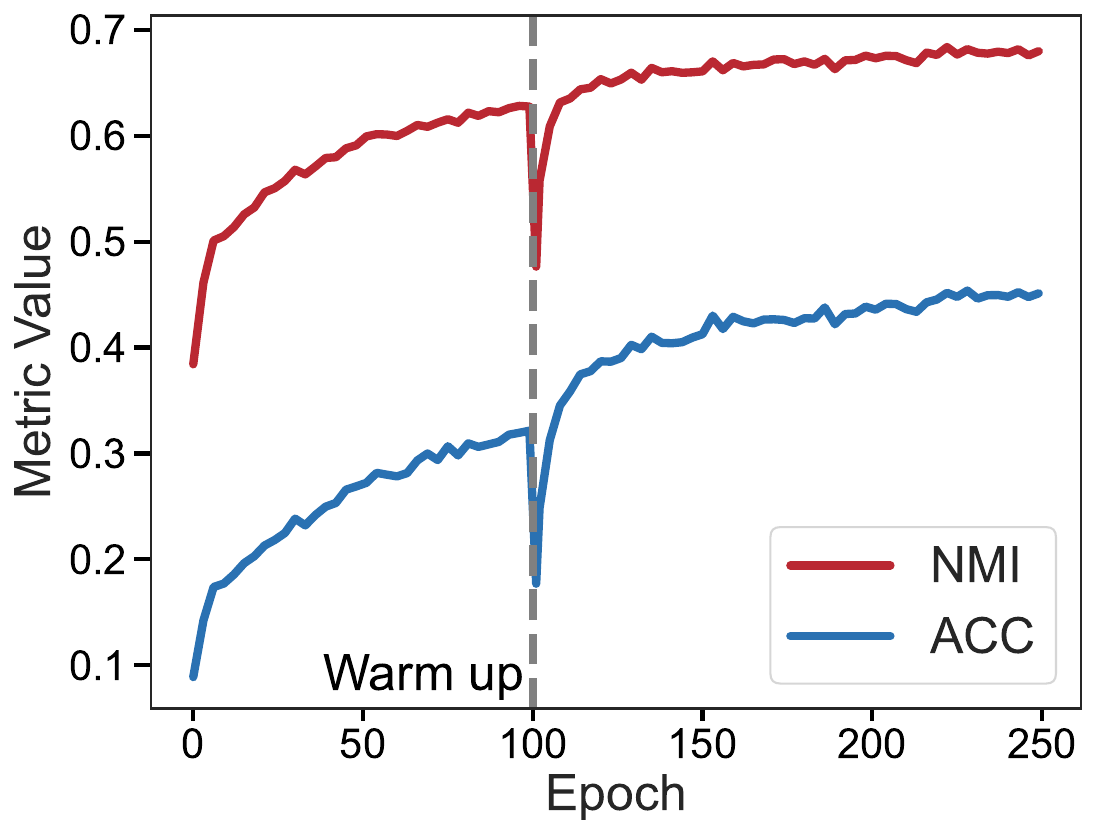} % Replace with actual image path
    \caption{Performance on Car}
    \label{fig:acc-b}
  \end{subfigure}
  \begin{subfigure}{0.24\linewidth}
    \centering
    \includegraphics[width=\linewidth]{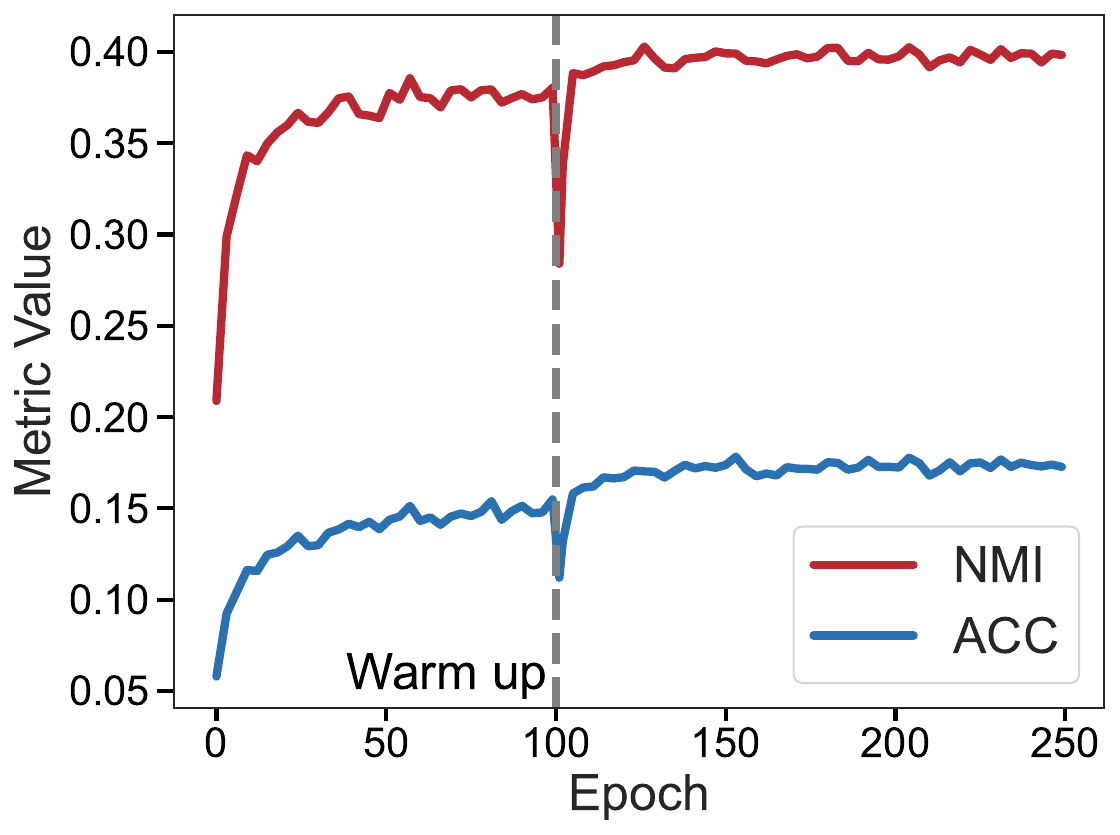} % Replace with actual image path
    \caption{Performance on Dog}
    \label{fig:acc-c}
  \end{subfigure}
  \begin{subfigure}{0.24\linewidth}
    \centering
    \includegraphics[width=\linewidth]{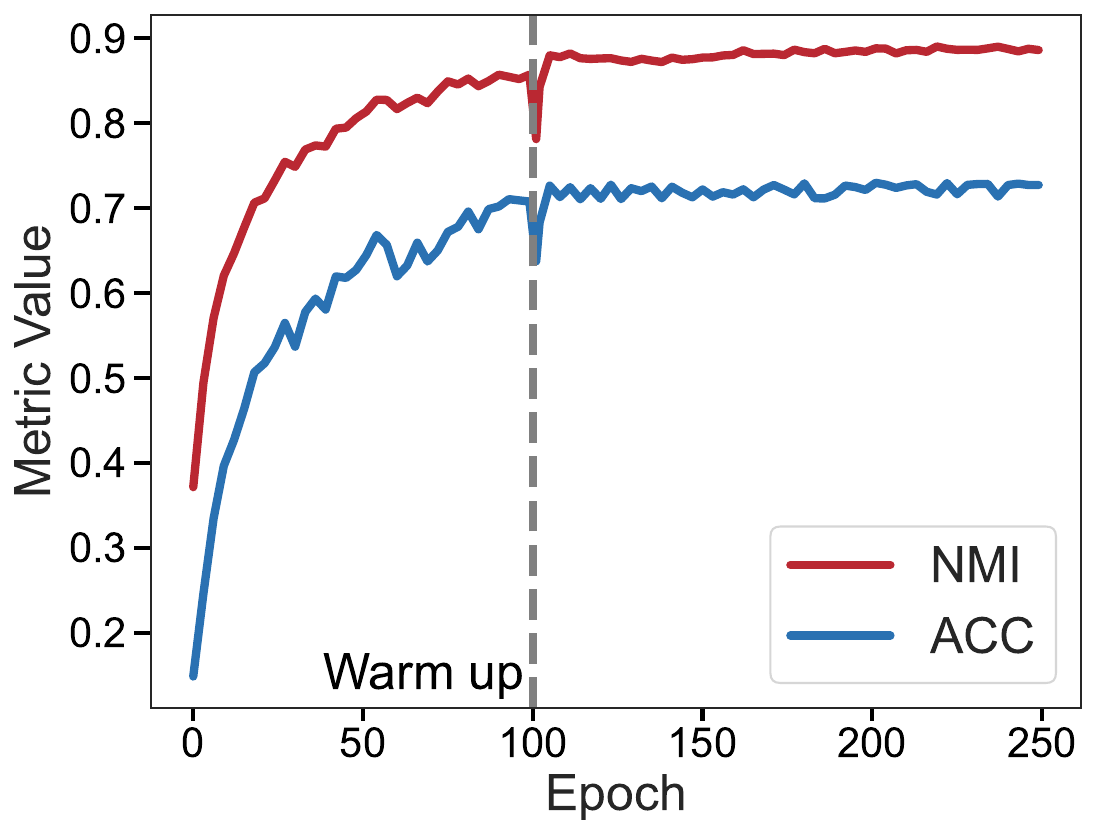} % Replace with actual image path
    \caption{Performance on Flower}
    \label{fig:acc-d}
  \end{subfigure}
  \caption{Clustering performance of DiFiC on four datasets across the training process, with ACC divided by 100 for simplicity. In the first 100 warm-up epochs, the performance refers to applying $k$-means on proxy words $S^*$. At epoch 100, $\mathcal{L}_{{CG}}$ and the clustering head $g(\cdot)$ are introduced. The sudden performance drop is due to the random initialization of $g(\cdot)$.}
  \label{fig:acc}
\end{figure*}

\begin{figure*}
  \centering
  \begin{subfigure}{0.24\linewidth}
    \centering
    \includegraphics[width=\linewidth]{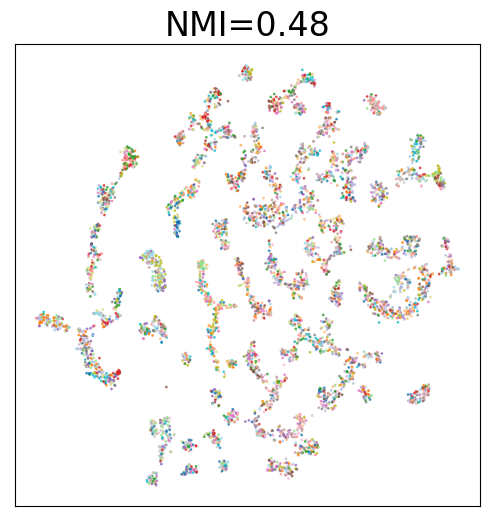} % Replace with actual image path
    \caption{SeCu}
    \label{fig:tsne-a}
  \end{subfigure}
  \begin{subfigure}{0.24\linewidth}
    \centering
    \includegraphics[width=\linewidth]{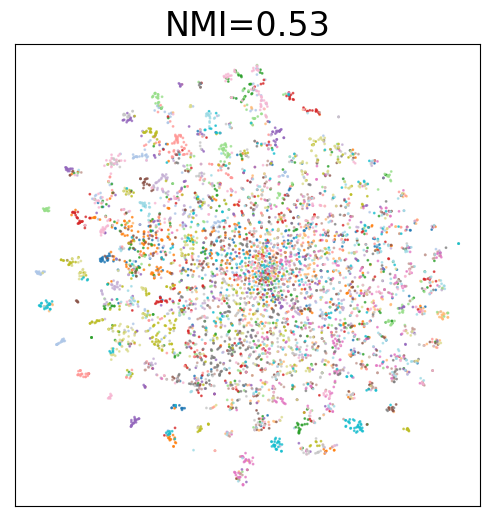} % Replace with actual image path
    \caption{C3-GAN}
    \label{fig:tsne-b}
  \end{subfigure}
  \begin{subfigure}{0.24\linewidth}
    \centering
    \includegraphics[width=\linewidth]{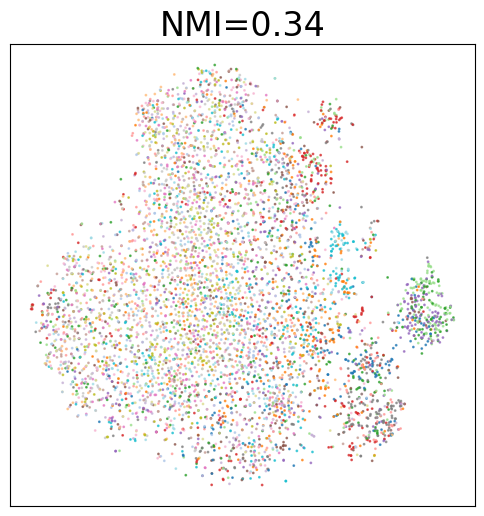} % Replace with actual image path
    \caption{Stable Diffusion}
    \label{fig:tsne-c}
  \end{subfigure}
  \begin{subfigure}{0.24\linewidth}
    \centering
    \includegraphics[width=\linewidth]{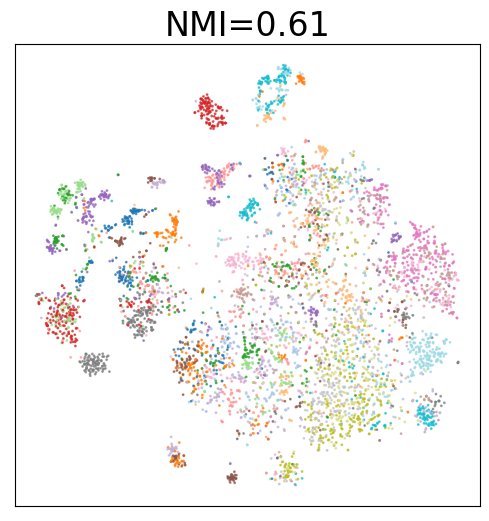} % Replace with actual image path
    \caption{DiFiC}
    \label{fig:tsne-d}
  \end{subfigure}
  \caption{Visualization of features learned by SeCu, C3-GAN, Stable Diffusion, and our DiFiC on CUB dataset, with the corresponding clustering NMI annotated at the top.}
  \vspace{-2mm}
  \label{fig:tsne}
\end{figure*}

In this section, we compare DiFiC with state-of-the-art clustering baselines on the four datasets, followed by convergence analysis and feature visualizations.

\begin{figure*}[t]
  \centering
   \includegraphics[width=\linewidth]{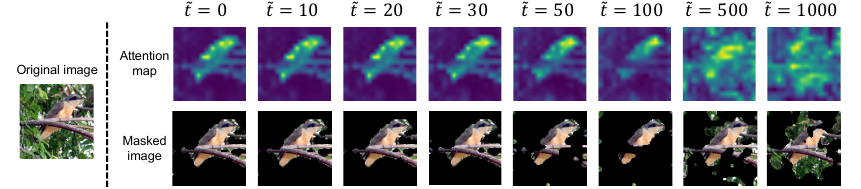}
   \caption{The cross-attention maps and the masked images of different $\tilde{t}$.}
   \vspace{-2mm}
   \label{fig:birdmask}
\end{figure*}

\subsubsection{Clustering Performance Comparison}

We compare DiFiC with 11 state-of-the-art baselines, including \textit{i}) discriminative methods: IIC \cite{iicdis4}, SimCLR \cite{simclrcl1}, MoCo \cite{mococl2}, SCAN \cite{scanpseudo3}, SeCu \cite{secudis8}; and \textit{ii}) generative methods: InfoGAN \cite{infogangen1}, FineGAN \cite{finegangen2}, MixNMatch \cite{mixnmatch}, OneGAN \cite{onegangen3}, Stable Diffusion \cite{sd}, and C3-GAN \cite{c3gangen4}. For FineGAN and MixNMatch, we benchmark the fully unsupervised version and semi-supervised variant that incorporates prior bounding boxes to locate the object during training. For C3-GAN, we report its standard and over-clustering results, following the configurations in the original paper. Notably, for fair comparisons, we recompute the clustering accuracy for over-clustering by finding a many-to-one mapping.

As shown in \cref{tab:sota}, the proposed DiFiC significantly outperforms state-of-the-art methods on all datasets. In particular, DiFiC achieves $9\%$ and $37\%$ ACC improvements on CUB and Car, compared with the second-best method. As DiFiC is built upon a pre-trained Stable Diffusion model, we benchmark the clustering performance by directly applying $k$-means on the image features ${z}_{\hat{t}}^b$ extracted by the UNet encoder. It turns out that DiFiC significantly outperforms such a naive baseline, demonstrating that the diffusion model itself cannot learn discriminative features, whereas our semantic distillation strategy can effectively capture the image semantics. Moreover, DiFiC outperforms the state-of-the-art discriminative method SeCu, even though the latter is boosted with an advanced pseudo-labeling strategy. Such a result could be attributed to the inherent limitation of data augmentations in capturing fine-grained characteristics. Lastly, DiFiC also surpasses existing generative clustering methods that focus on learning latent image features, demonstrating the superiority of distilling semantics into text conditions, as proposed in this work.

\subsubsection{Convergence Analysis}

To evaluate the convergence of DiFiC, we plot its clustering performance with respect to training epochs in \cref{fig:acc}. As can be seen, the performance of DiFiC continuously and steadily increases across the training process. The only exception is at epoch 100 when the randomly initialized clustering head is introduced. On all four datasets, DiFiC archives a good convergence within 250 training epochs.

\subsubsection{Feature Visualization}

To provide an intuitive understanding of the clustering results, we visualize the CUB features extracted by the state-of-the-art discriminative method SeCu, generative method C3-GAN, the pre-trained Stable Diffusion model, and our DiFiC in \cref{fig:tsne}. As shown, the latent image features extracted by SeCu and C3-GAN display severe overlap across different classes. In contrast, the textual semantics distilled by our DiFiC form more compact and separated clusters. In other words, compared with directly learning the image representation, deducing text conditions used for image generation, as proposed in this work, is more effective in mining fine-grained image semantics. Moreover, \cref{fig:tsne-c} shows that the Stable Diffusion model itself cannot extract discriminative image features, which demonstrates that the success of our DiFiC lies in the semantic distillation paradigm, instead of the feature extraction ability of diffusion models.

\subsection{Ablation Study}

To better understand the effectiveness of the proposed DiFiC, we conduct ablation studies on the three modules and the results are shown in \cref{tab:abl}. From the results, one could arrive at three conclusions: \textit{i}) by deducing text conditions used for image generation, image semantics could be successfully distilled into the proxy word, significantly improving the feature discriminability compared with the latent image features extracted by the diffusion model; \textit{ii}) applying the mask to regularize the diffusion target reduces abundant background information and helps the distilled semantics focus on the main object in images, thus improving clustering performance; and \textit{iii}) by further leveraging neighborhood similarly, DiFiC distills more clustering-favorable textual semantics, leading to the best performance.

\begin{table}[t]
  \caption{Clustering performance with different combinations of the three modules on CUB and Car, where ``SD'', ``OC'', and ``CG'' denote the semantic distillation, object concentration, and cluster guidance module, respectively.}
  \label{tab:abl}
  \centering
  \begin{tabular}{ccccccc}
    \toprule
    \multirow{2}{*}{SD} & \multirow{2}{*}{OC} & \multirow{2}{*}{CG} & \multicolumn{2}{c}{CUB} & \multicolumn{2}{c}{Car} \\
    \cmidrule(lr){4-5} \cmidrule(lr){6-7}
     & & & ACC & NMI & ACC & NMI \\
    \midrule
    \multicolumn{3}{c}{\text{Stable Diffusion}} & 7.6 & 0.34 & 9.1 & 0.37 \\
    \midrule
    \checkmark & & & 22.8 & 0.52 & 43.7 & 0.65 \\
    \checkmark & \checkmark & & 26.1 & 0.56 & 45.5 & 0.66 \\
    \checkmark & & \checkmark & 26.8 & 0.57 & 41.8 & 0.65 \\
    \checkmark & \checkmark & \checkmark & \textbf{31.7} & \textbf{0.61} & \textbf{47.2} & \textbf{0.68}\\
    \bottomrule
  \end{tabular}
  \vspace{-2mm}
\end{table}

\subsection{Parameter Analyses}
We analyze the influence of two key parameters in the proposed DiFiC, the timesteps $t$ used for semantic distillation and the timestep $\tilde{t}$ used for generating the object mask.

\subsubsection{Timesteps \texorpdfstring{$t$}{t} for Semantic Distillation}

Recall that when distilling the textual semantics, we propose emphasizing small diffusion timesteps to better capture fine-grained details through a weighted timestep sampling scheme in \cref{eq:wt}. For further investigation, we test four sampling schemes to assess how different diffusion timesteps influence the clustering performance. As shown in \cref{tab:wt}, random sampling from timesteps 0--500 achieves better clustering performance than from timesteps 500--1000, suggesting that low-level features generated at smaller diffusion timesteps are more favorable for fine-grained clustering. Notably, compared with roughly ignoring the large timesteps, sampling from all timesteps in a weighted manner yields better performance. Such a result naturally makes sense as the richer low- or high-level features at different diffusion stages are only relative instead of absolute.

\begin{table}[!h]
  \caption{Influence of different timestep sampling schemes in semantic distillation on CUB and Car.}
  \label{tab:wt}
  \centering
  \begin{tabular}{rcccc}
    \toprule
    \multirow{2}{*}{Timesteps} & \multicolumn{2}{c}{CUB} & \multicolumn{2}{c}{Car} \\
    \cmidrule(lr){2-3} \cmidrule(lr){4-5}
     & ACC & NMI & ACC & NMI \\
    \midrule
     0--500 (Random) & 29.8 & 0.60 & 44.2 & 0.66 \\
     500--1000 (Random) & 21.3 & 0.52 & 33.8 & 0.61 \\
     0--1000 (Random) & 27.2 & 0.58 & 45.1 & 0.67 \\
     0--1000 (Weighted) & \textbf{31.7} & \textbf{0.61} & \textbf{47.2} & \textbf{0.68}\\
    \bottomrule
  \end{tabular}
\end{table}

\begin{figure}
  \centering
   \includegraphics[width=0.95\linewidth]{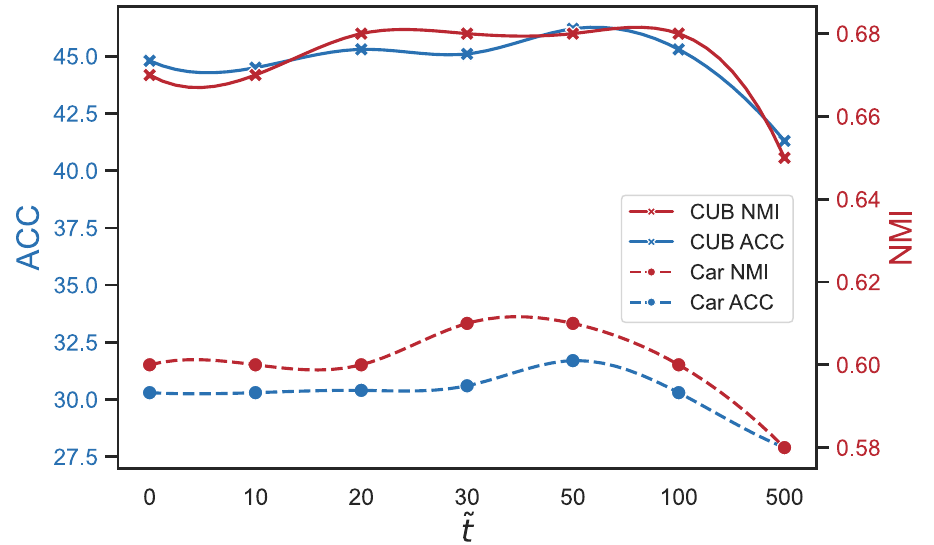}
   \caption{Influences of $\tilde{t}$ for mask generation on CUB and Car.}
   \vspace{-1mm}
   \label{fig:maskp}
\end{figure}

\subsubsection{Timestep \texorpdfstring{$\tilde{t}$}{t} for Mask Generation}

In the object concentration module, we use a fixed timestep $\tilde{t}$ when calculating the object mask via \cref{eq:atten}. Here, we try different choices of $\tilde{t}$ to study its influence on locating the object in images. As illustrated in \cref{fig:maskp}, the best performance is achieved for $\tilde{t}$ around 50. When the diffusion timestep is too large, the input appropriates to Gaussian noise, leading to a scattered attention map that fails to capture the main object as shown in \cref{fig:birdmask}. On the contrary, when the timestep is too small, the attention map would mistakenly recognize some high-frequency backgrounds (\textit{e.g.}, the branch in \cref{fig:birdmask}) as part of the main object, leading to inferior performance. Notably, the best choice of $\tilde{t}$ is consistent across the two datasets.

\section{Conclusion}
In this paper, we propose a novel fine-grained clustering method DiFiC, based on the conditional diffusion model. Unlike most existing clustering approaches that focus on extracting representative image features, DiFiC distills image semantics by deducing the textual condition used for image generation. Enhanced by the proposed object concentration and cluster guidance strategies, DiFiC effectively addresses the semantic disruption and redundancy problem suffered by existing discriminative and generative clustering methods. On four fine-grained image clustering datasets, DiFiC demonstrates a precise capture of subtle semantic differences across classes, achieving state-of-the-art performance compared with existing baselines. For future research, on the one hand, DiFiC demonstrates that models designed for other tasks could serve as effective external knowledge to facilitate clustering. This opens intriguing possibilities for incorporating broader, and even seemingly unrelated, types of external knowledge to guide clustering \cite{extra}. On the other hand, this work reveals the untapped potential of diffusion models in clustering tasks. We hope the success of DiFiC could inspire further exploration of diffusion models across a wider range of applications beyond image generation.

{
    \small
    \bibliographystyle{ieeenat_fullname}
    \bibliography{main}
}


\end{document}